\documentclass[letterpaper]{article} 
\usepackage{aaai2026}  
\usepackage{times}  
\usepackage{helvet}  
\usepackage{courier}  
\usepackage[hyphens]{url}  
\usepackage{graphicx} 
\usepackage{natbib}  
\usepackage{caption} 
\usepackage{booktabs}
\usepackage{multirow}
\usepackage{amsmath}
\usepackage{amssymb}
\usepackage{amsfonts}
\usepackage{subcaption}
\usepackage{fancyhdr}

\usepackage{algorithm}
\usepackage{algorithmic}

\usepackage{newfloat}
\usepackage{listings}
\DeclareCaptionStyle{ruled}{labelfont=normalfont,labelsep=colon,strut=off} 
\floatstyle{ruled}
\newfloat{listing}{tb}{lst}{}
\floatname{listing}{Listing}
\newcommand{\methodabbr}{AuthSig}
\title{\methodabbr: Safeguarding Scanned Signatures Against Unauthorized Reuse \\ in Paperless Workflows}
\author{
    Ruiqiang Zhang\textsuperscript{\rm 1},
    Zehua Ma\textsuperscript{\rm 1}\equalcontrib,
    Guanjie Wang\textsuperscript{\rm 1},
    Chang Liu\textsuperscript{\rm 1},
    Hengyi Wang\textsuperscript{\rm 1},
    Weiming Zhang\textsuperscript{\rm 1}\equalcontrib\\
}
\affiliations{
    \textsuperscript{\rm 1}Anhui Province Key Laboratory of Digital Security, University of Science and Technology of China\\

}

\usepackage{bibentry}

\begin{document}
\captionsetup[figure]{name={Fig.},labelsep=period,justification=raggedright,font={small}}
\maketitle

\begin{abstract}
With the deepening trend of paperless workflows, signatures as a means of identity authentication are gradually shifting from traditional ink-on-paper to electronic formats. Despite the availability of dynamic pressure-sensitive and PKI-based digital signatures, static scanned signatures remain prevalent in practice due to their convenience. However, these static images, having almost lost their authentication attributes, cannot be reliably verified and are vulnerable to malicious copying and reuse. To address these issues, we propose \textbf{\methodabbr}, a novel static electronic signature framework based on generative models and watermark, which binds authentication information to the signature image. Leveraging the human visual system’s insensitivity to subtle style variations, \methodabbr{} finely modulates style embeddings during generation to implicitly encode watermark bits---enforcing a One Signature, One Use policy. To overcome the scarcity of handwritten signature data and the limitations of traditional augmentation methods, we introduce a keypoint-driven data augmentation strategy that effectively enhances style diversity to support robust watermark embedding. Experimental results show that \methodabbr{} achieves over 98\% extraction accuracy under both digital--domain distortions and signature--specific degradations, and remains effective even in print-scan scenarios.

\end{abstract}

\section{Introduction}
Handwritten signatures have long served as a trusted means of identity authentication in domains such as financial transactions and legal documentation, and they remain integral to contemporary social credit systems. However, in the wake of digital transformation and the rise of paperless workflows \cite{sellen2003myth}, electronic signatures are becoming the preferred choice over conventional ink-on-paper signatures in many contexts. According to the E-SIGN Act \cite{esign:2000}, an electronic signature is ``an electronic sound, symbol, or process” that is attached to or logically associated with a contract or other record and executed by a person with the intent to sign. Common implementations include digital certificates, email confirmations, and password‑protected workflows. Consequently, numerous online signature platforms have emerged, such as DocuSign, Adobe Sign, and Signaturit \cite{alexander2019easy}, 
\begin{figure}[tbp]
    \centering
    \includegraphics[width=0.8\columnwidth]{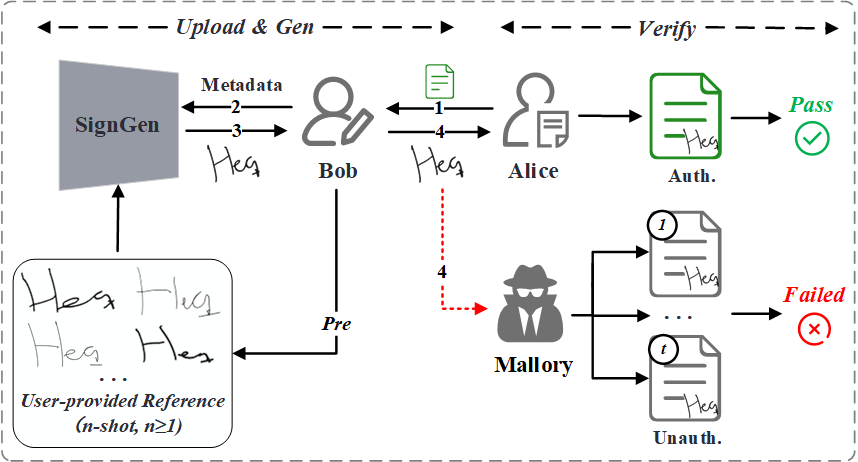}
    \caption{System overview. Bob pre-registers n signature samples. For Alice's request, the system generates a metadata-bound signature valid only for this verification. Reuse with mismatched documents will fail authentication. The numbered arrows indicate the chronological sequence of operations.}
    \label{fig:system_pipeline}
\end{figure}
which fall into three main paradigms: dynamic signatures \cite{okawa2021time,okawa2020online}, which leverage dedicated hardware to capture writing dynamics (e.g., pen pressure, velocity)  as used by platforms like Signaturit to validate authenticity via behavioral biometrics; Public Key Infrastructure (PKI)-based digital signatures, where services such as DocuSign and Adobe Sign employ certificates issued by Certificate Authorities to bind user identity and ensure document integrity; and static scanned-signature images, where users capture an image of their handwritten signature via scanning or photography and paste it into a digital document.

Despite their technical merits, current digital-signature methods generally suffer from a detachment between signature appearance and authentication attributes. In the first two paradigms, dynamic signatures rely on biometric handwriting features and PKI digital signatures rely on cryptographic algorithms; in both cases, the authentication process depends on externally derived attributes (dynamic features or PKI certificates) rather than on the visual presentation of the signature. Furthermore, these methods are often encumbered by expensive proprietary hardware and poor cross-platform compatibility. The third paradigm, static scanned signature images, further degrades security. Verification of these images lacks reliable biometric support, and their reusability severs the essential link between the signing act and the specific document content. Nonetheless, their operational simplicity and visual clarity sustain their prevalence in low-risk contexts (e.g., internal approvals, non-binding accords).
Yet, if misappropriated, a static signature image can be illicitly reused to forge documents \cite{LPLFinancial2023AWC}, with no post-hoc mechanism to conclusively attribute authorship, thereby exposing organizations and individuals to dispute and liability.

To overcome these limitations, we pursue a watermark-based static electronic signature framework that enables signature-verification binding and natural sign-authenticate workflows. As illustrated in Fig. \ref{fig:system_pipeline}, upon Alice’s authorization request, Bob encodes metadata (document, signatory, expiration, etc.) into a watermark bitstream. He then combines the watermark with his pre-registered signature samples to generate a one-time-use watermarked signature image, which is delivered to Alice. During verification, the watermark is extracted and validated, and associated metadata are queried to enable granular checks (e.g. document binding, expiration). When an attacker illicitly reuses the signature image, it won't pass system verification, enforcing the ``One Signature, One Use" mechanism. 

However, the widely used watermarking paradigms, primarily designed for natural images with post-processing methods, are ill-suited to the target framework, as they often cause visible artifacts in smooth regions and lack robustness to typical document transformations when applied to signature images. Thus, we propose a generative signature watermarking framework that fuses watermark with dynamic signature style features and employs a generative reconstruction mechanism for implicit embedding, establishing a strong link between authorization and the signature. Furthermore, to overcome the limitation posed by the small scale of existing handwritten signature datasets on generative model training, a keypoint-driven data augmentation strategy is designed. For embedding via dynamic features without altering the signature’s character structure, we employ supervised contrastive learning to extract signer-invariant content features and jointly train a style encoder with our generator for fine-grained, implicit modeling of signature style.

In summary, the contributions of this paper are as follows: 
\begin{itemize}
    \item We propose \methodabbr{}, a electronic signature framework that leverages watermarking to equip static signatures with verifiability and document-binding properties. This enables an intuitive, natural signing-and-verification process closely resembling traditional ink signatures.
    \item We devise a keypoint-driven data augmentation strategy tailored for glyph data. This approach effectively mitigates the scarcity of training data by enriching signature sample diversity while meticulously preserving the topological structure of the handwriting.
    \item Based on a generative model, we constructed a end-to-end watermarking architecture: it fuses watermark information with signature style features in the feature space and leverages generative reconstruction for implicit watermarking. This approach delivers strong robustness against both common digital-domain distortions and those specific to signature scenarios, even including print-scan process.
\end{itemize}

\section{Related Work}
\subsection{Handwriting Imitation}
Handwriting synthesis has progressed via GANs \cite{goodfellow2020generative} and diffusion models \cite{sohl2015deep,ho2020denoising}. HiGAN+ \cite{gan2022higan+} disentangles content and style for realistic generation but is constrained by training bias. Diffusion-based models—WordStylist \cite{nikolaidou2023wordstylist}, CTIDiffusion \cite{zhu2023conditional}, and One-DM \cite{dai2024one}—enable cross-lingual and few-shot synthesis with improved fidelity. Recent works like DiffusionPen \cite{nikolaidou2024diffusionpen} and StylusAI \cite{riaz2024stylusai} offer fine-grained control and stylistic generalization. However, these models rely on labeled datasets (e.g., IAM \cite{marti2002iam}) and assume consistent writer styles, while signature data often lack content labels and exhibit high intra-writer variability, challenging generalization.

\subsection{Signature Image Verification}
Offline signature verification has long been approached as a biometric classification task, utilizing methods such as SVMs \cite{pal2013two}, Siamese networks \cite{dey2017signet,avola2021r,chattopadhyay2022surds}, and graph-based models \cite{maergner2019graph} to learn writer-specific features from static signature images. While these techniques achieve strong performance on benchmarks like CEDAR, they are often brittle under cross-domain shifts and high intra-writer variability. More critically, most existing approaches focus exclusively on visual or biometric similarity, overlooking the intent of the signature and the potential for unauthorized reuse in digital contexts. Digital watermarking offers a potential solution by enabling content binding and usage tracking for static images, yet current methods—originally designed for documents or natural images—face significant challenges when applied to handwritten signatures. Document-based techniques, for instance, rely on structured text layouts \cite{yang2023autostegafont}, which are absent in free-form handwriting. Meanwhile, image watermarking approaches exploit high-frequency textures, but signature images concentrate meaningful pixels within continuous strokes—only about 14\% of the total pixels—making such methods less effective. For example, HiDDeN \cite{zhu2018hidden} and CIN \cite{ma2022towards} use global high-frequency embeddings that often place watermarks in background regions, reducing imperceptibility. These limitations underscore the difficulty of directly applying existing watermarking techniques to the domain of signature image verification.

\begin{figure*}[htbp]
    \centering
    \includegraphics[width=0.93\textwidth]{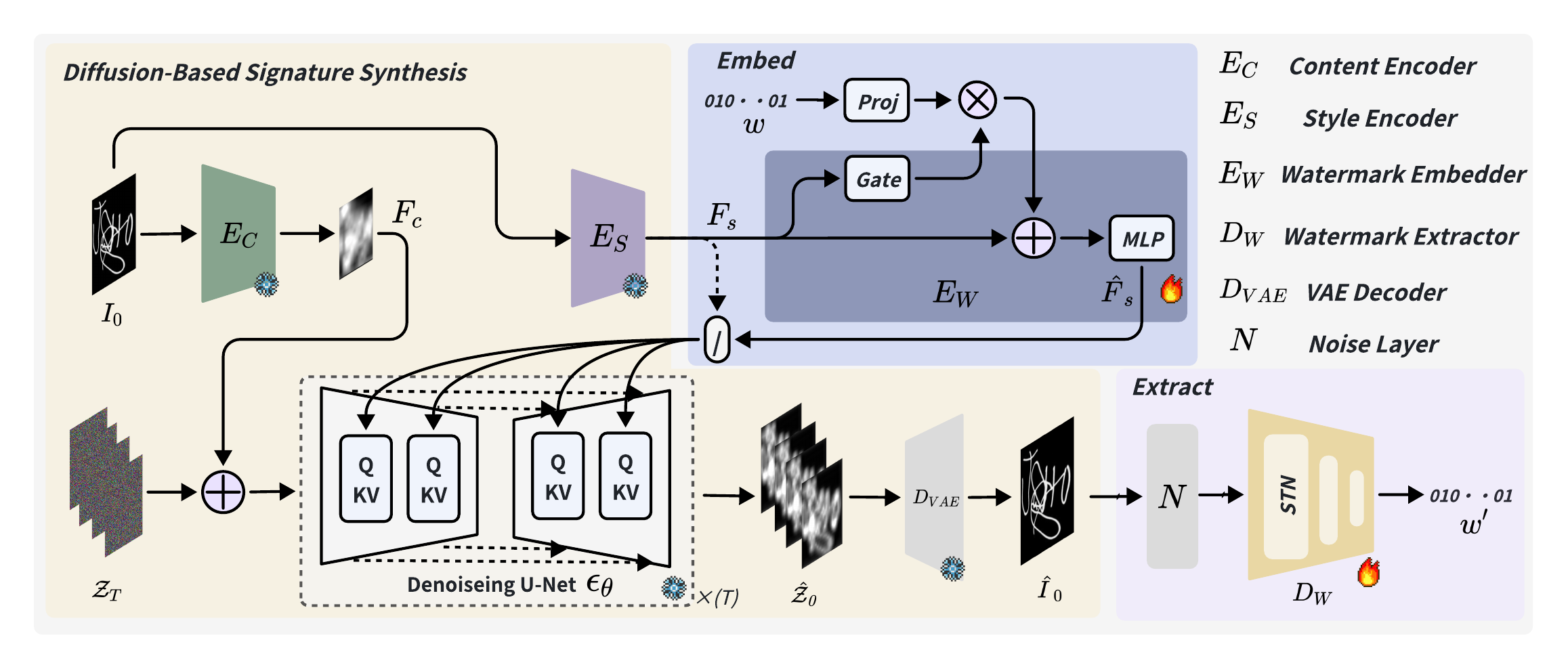}
    \caption{The training process of the proposed watermarking model. \textbf{Embedding}: Content feature $F_c$ is concatenated with noise latent $z_T$, while style feature $F_s$ is fused with watermark bits $w$ via a gated-residual block; this conditioned input guides the diffusion model to generate the watermarked signature. \textbf{Extraction}: Extractor $E_D$ employs STN for geometric alignment and a multi-layer MLP to recover $w$. A distortion layer $N$ enhances training robustness.}
    \label{fig:method_pipeline}
\end{figure*}

\section{Method}
\methodabbr{} comprises three sequential stages. First, we abstract each signature’s topology through skeleton extraction and keypoint filtering, followed by TPS–based data augmentation to simulate realistic handwriting variations. Second, as shown in Fig. \ref{fig:method_pipeline}, we employ supervised contrastive learning to derive signer-invariant content features and architect a latent-space diffusion model that integrates style features and structural cues for high-fidelity signature reconstruction. Third, we introduce a diffusion-based implicit watermark framework that injects watermark into style features via an adaptive gated residual fusion mechanism, ensuring robust, imperceptible watermarking within the generated signature.

\subsection{Keypoint-Based Structural Augmentation}
\begin{figure}[thbp]
    \centering
    \includegraphics[width=1\columnwidth ]{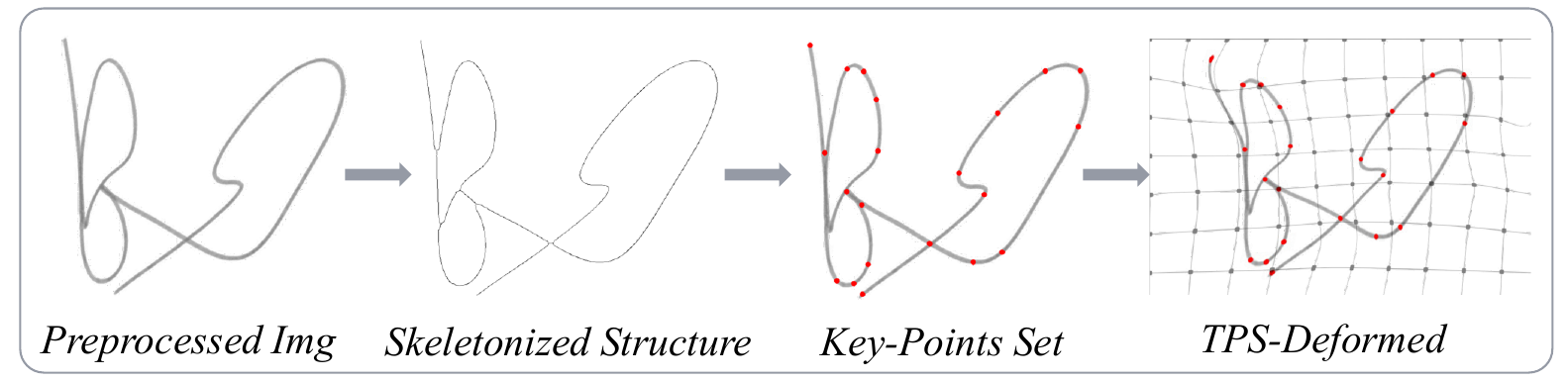}  
    \caption{Structural augmentation pipeline: the input signature is skeletonized, keypoints are extracted, and a Thin Plate Spline (TPS) transformation is applied for topologically consistent deformation.}
    \label{fig:data_aug} 
\end{figure}

Considering that existing image-augmentation strategies applied directly to signature images easily cause stroke discontinuities and overlaps. We abstract the signature from its spatial pixel form to a more essential topological structure using signature skeletonization and a keypoint filtering mechanism. By subsequently applying a Thin Plate Spline (TPS) transformation to the keypoints, as shown in Fig. \ref{fig:data_aug}, we ensure that the resulting augmented samples both retain the structural integrity of the original signature and effectively model the natural variability of authentic signatures.

Specifically, for an input signature image $I \in \mathbb{R}^{H \times W \times 3}$, we first convert it to grayscale ($I_{\text{gray}} \in \mathbb{R}^{H \times W}$), binarize it ($I_{\text{bin}} \in \{0,1\}^{H \times W}$), and apply morphological opening to obtain the preprocessed signature $I_{\text{pre}}$.
Then, the Zhang-Suen thinning algorithm is adopted for skeleton extraction on $I_{\text{pre}}$, which provides a set of skeleton points $\mathcal{S} \subset \mathbb{Z}^2$. From this set, we extract a simplified set of critical points $\mathcal{P}$ using the Ramer-Douglas-Peucker (RDP) algorithm:
\begin{equation}
\mathcal{P} = \text{RDP}(\mathcal{S}, \varepsilon) = \{(x_1, y_1), (x_2, y_2), \dots, (x_k, y_k)\}
\label{eq:rdp}
\end{equation}
where $\varepsilon$ governs the density of the extracted points, and each $(x,y)$ denotes the coordinates of a skeleton keypoint.

Finally, the signature image $I$ is warped using a Thin Plate Spline (TPS) transformation. We randomly sample a subset of the keypoints:
\begin{equation}
\mathcal{C} = \left\{ (x_1, y_1), (x_2, y_2), \dots, (x_n, y_n) \right\} \subset \mathcal{P}
\label{eq:control_points}
\end{equation}
to serve as control points. Each control point $(x_i, y_i) \in \mathcal{C}$ is perturbed by a random displacement from a truncated Gaussian distribution:
$\delta_{x_i}, \delta_{y_i} \sim \mathcal{N}_{\text{trunc}}(0, \sigma^2, [-k\sigma,k\sigma])$
with $\sigma = 1$ and a truncation parameter $k \in \mathbb{R}^+$. The destination coordinates $(x_i',y_i')$ are clamped within the image bounds:
\begin{equation}
\begin{aligned}
x_i' &= \max(0, \min(W, x_i + \delta_{x_i})) \\
y_i' &= \max(0, \min(H, y_i + \delta_{y_i}))
\end{aligned}
\label{eq:clipping}
\end{equation}
In this manner, the signature image undergoes a smooth, non-rigid deformation that preserves the overall topological structure of the strokes. This process effectively simulates variations in writing dynamics, such as angle and ligature, generating highly realistic samples. 

\subsection{Signature Synthesis Network}
\subsubsection{Feature Decoupling}
Let the signature dataset be $\mathcal{D} = \left\{ (I_k, id_k)\right\}_{k=1}^N$, where $I_k \in \mathbb{R}^{H \times W \times C}$ is an input image and $id_k$ is the corresponding signer identity label. We hypothesize that a signature's feature representation can be decomposed into two components: a \textit{content feature} and a \textit{style feature}. The former captures semantic information such as character structure and stroke composition, which should remain consistent across all the user's signatures; the latter reflects idiosyncratic signing dynamics and variations unique to the writer. Following this hypothesis, the set of features for all $m$ samples belonging to a single signer $id_i$ can be formally represented as:$\left\{ C^k_{\mathbf{s}_1}, C^k_{\mathbf{s}_2}, \ldots, C^k_{\mathbf{s}_ j} \right\} \subset \mathbb{R}^{d}$. In this notation, $C^k$ denotes the content feature of the $k$-th signer, while $\mathbf{s}_j$ denotes the instance-specific style feature of sample $j$. As existing datasets lack explicit annotations for signature content, we propose to learn a content feature encoder, $E_c$, using a Supervised Contrastive Learning framework.

Given a mini-batch of $B$ samples drawn from the signature dataset $\mathcal{D}$, let
\begin{equation}
    c_k =\frac{\textit{Proj}\bigl(E_c(I_k)\bigr)}{\bigl\|\textit{Proj}\bigl(E_c(I_k)\bigr)\bigr\|} \in\mathbb{R}^d
\end{equation}
denote the normalized, projected content embedding of the input image $I_k$. Here $E_c$ is the content-feature encoder and $Proj(\cdot)$ is the learnable projection head onto the unit hypersphere. For $k\in\{1,\dots,B\}$, define $P(k)=\bigl\{\,p\in A(k) \mid id_p=id_k\bigr\}$ as the set of positive examples sharing the same signer label $id_k$, and $A(k)=\{1,\dots,B\}\setminus\{k\}$ as the set of all other indices in the batch. Therefore, the training objective for $E_c$ is
\begin{equation}
\mathcal{L}_{SupCon} = \sum_{k=1}^{B} \frac{-1}{|P(k)|} \sum_{p \in P(k)} \log \frac{\exp(c_k \cdot c_p / \tau)}{\sum_{a \in A(k)} \exp(c_k \cdot c_a / \tau)}
\end{equation}
where $\tau$ is a temperature hyperparameter that controls the concentration of similarity scores. 
By minimizing $\mathcal{L}_{SupCon}$, it promotes encoder $E_c$ to learn feature representations that possess high cohesion for same-identity signatures and high separability for different-identity signatures, compelling $E_c$ to extract content features related to character structure and semantic information.

\subsubsection{Diffusion-Based Signature Synthesis}
Diffusion models typically operate in a low-dimensional latent space to ensure computational efficiency, employing a Variational Autoencoder (VAE) to map images from pixel space to this latent representation \cite{rombach2022high}. However, standard pre-trained VAEs often fail to preserve the fine-grained details of signatures. This stems from the under-representation of intricate glyph data in their original training corpora, leading to significant information loss during reconstruction. To address this, we perform domain-specific fine-tuning on the VAE before using it in our diffusion pipeline. The process is guided by a composite objective function $\mathcal{L}_{\text{VAE}}$, which combines three distinct loss terms:
\begin{equation}
    \mathcal{L}_{\text{VAE}} = \mathcal{L}_{\text{rec}} + \mathcal{L}_{\text{masked\_rec}} + \lambda \mathcal{L}_{\text{KL}}
\end{equation}
Here, $\mathcal{L}_{\text{rec}}$ penalizes pixel-wise reconstruction errors, $\mathcal{L}_{\text{masked\_rec}}$ emphasizes regions marked by a binary mask $m$, and $\mathcal{L}_{\text{KL}}$ regularizes the latent space. This targeted tuning enables the VAE to better preserve critical stroke-level details essential for high-fidelity diffusion generation.

Building upon this optimized latent representation, our proposed algorithm for signature image reconstruction leverages a Denoising Diffusion Probabilistic Model \cite{ho2020denoising}. To achieve high-fidelity and identity-consistent reconstruction, two distinct types of conditional features are incorporated to guide this denoising process. As shown in Fig.\ref{fig:method_pipeline}, $F_s$ is the style feature of the signature image, extracted by the style encoder $E_s$ to capture the signer’s dynamic and individualistic characteristics and integrated via a cross‑attention mechanism, whereas $F_c$, obtained from the content encoder $E_c$, encodes the structural information of the signature strokes and is fused with the timestep embedding vector $t_{\mathrm{emb}}$ at each denoising block. Accordingly, the training objective at this stage is defined as
\begin{equation}
\begin{split}
    \mathcal{L}_{\text{denoise}}(\theta) ={} & \mathbb{E}_{\mathcal{Z}_0 \sim q(\mathcal{Z}_0), t \sim \mathcal{U}[1, T], \epsilon \sim \mathcal{N}(0, \mathbf{I})} \\
    & \quad \left[ \left\| \epsilon - \epsilon_\theta(\mathcal{Z}_t, t, F_s, F_c) \right\|^2 \right]
\end{split}
\end{equation}
Here, $\mathcal{Z}_t$ is obtained via the reparameterization trick. During inference, we initiate the reverse diffusion process from a pure noise sample $\mathcal{Z}_T \sim \mathcal{N}(0, \mathbf{I})$ and iteratively apply the reverse steps. To accelerate this sampling, we employ Denoising Diffusion Implicit Models (DDIM) \cite{song2020denoising}. Once the denoised latent representation $\hat{\mathcal{Z}_0}$ is obtained, a VAE decoder $\mathcal{D}$ reconstructs the final signature image $\hat{I_0} = \mathcal{D}(\hat{\mathcal{Z}_0})$. This framework balances computational complexity with generation quality by performing diffusion in the latent space, while the integration of conditional features critically ensures both the identity consistency and structural accuracy of the generated signatures.

\subsection{Watermark Embedding \& Extraction}
\begin{figure}[thbp]
    \centering
    \includegraphics[width=0.9\columnwidth]{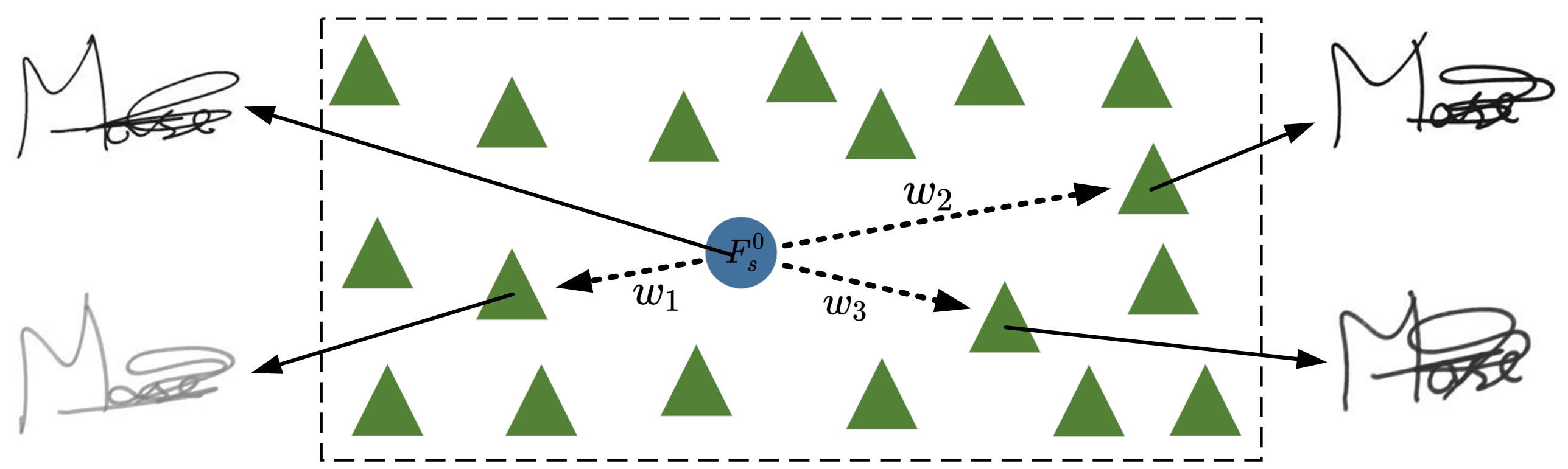}
    \caption{Schematic diagram of watermark embedding process.}
    \label{fig:watermarkemb}
\end{figure}

To overcome the challenges of quality degradation and limited capacity associated with directly embedding bit-level information into signature images, we introduce a diffusion-based framework for robust bit embedding and extraction, as illustrated in Fig.~\ref{fig:watermarkemb}. By fusing the binary information with semantic image features and injecting it as guidance into the diffusion model, our method achieves a balanced trade-off among embedding capacity, robustness, and visual fidelity.

The process begins by decomposing the input image $ I $ into content features $ F_c $ and style features $ F_s $ using content encoder $ E_c $ and style encoder $ E_s $. $ F_c $ captures intrinsic signature characteristics (e.g., strokes and structure), while \( F_s \) reflects signer-specific stylistic variations. In practice, $F_c$ can be extracted from any signature image of the signer, and the reference style feature $F_s^0$ is computed from a few prior signature samples.
we embed watermark $ w $ (where $ w_i \in \{0,1\} $) into $ F_s^0 $ to preserve semantic via a gated residual fusion in watermark embedder $E_W$:
\begin{equation}
    F_s^\prime = F_s^0 + g \cdot \textit{Proj}(w)
\end{equation}
where $\textit{Proj}(\cdot)$ is a projection layer and $g$ is a learned gating parameter. Then \( F_s^\prime \) serves as the conditioning input for the diffusion model to reconstruct an image \( \hat{I} \) that carries the embedded information.

The extraction process reverses this pipeline. The watermark extractor $D_W$ first applies a Spatial Transformer Network (STN) \cite{jaderberg2015spatial} to correct the input image $\hat{I}$. It then recovers the fused feature $\hat{F}_s^\prime$, concatenates it with a pre-defined $F_s^0$, which serves as a signer-specific prior. Unlike generic watermark verification, signature verification assumes a known identity, making it natural and feasible to access the corresponding reference style feature. The combined features are fed into an MLP to predict the embedded bits:
\begin{equation}
    w' = \textit{MLP}([\hat{F}_s^\prime \oplus F_s^0])
\end{equation}
During training, we freeze the parameters of \( E_c \), \( E_s \), the denoising model \( \epsilon_{\theta} \), and the VAE, optimizing only \( E_W \) and \( D_W \). The total loss is defined as:
\begin{equation}
    \mathcal{L}_{\text{total}} = \lambda_1 \mathcal{L}_{\text{secret}} + \lambda_2 \mathcal{L}_{\text{recover}} + \lambda_3 \mathcal{L}_{\text{style}} + \lambda_4 \mathcal{L}_{\text{content}}
\end{equation}
The individual losses are formulated as:
\begin{equation}
\mathcal{L}_{\text{secret}} = \textit{BCE}(w, w'),\quad \mathcal{L}_{\text{recover}} = \textit{MSE}(F_s^0, \hat{F}_s') 
\end{equation}
\begin{equation}
\mathcal{L}_{\text{style}} = \textit{MSE}(F_s^0,\,F_s') + \lambda_t \cdot \textit{ReLU}(m - ||F_s^0-F_s'||_2)
\end{equation}

Finally, to maintain the structural integrity of the signature, $\mathcal{L}_{\text{content}}$ is formulated using the Contextual Loss \cite{mechrez2018contextual}:
\begin{equation}
    \mathcal{L}_{\text{content}} = -\log\left(\text{CX}\left(E_c(I), E_c(\hat{I})\right)\right)
\end{equation}
where $\text{CX}(\cdot, \cdot)$ measures the similarity between feature sets. 

\section{Experiments} \label{sec:exper}
\subsection{Implementation Details}
We train our model on the GPDS 1–150 dataset \cite{vargas2007off}, which comprises 150 signers with 54 signature images each. We randomly allocate 120 signers’ signatures to the training set and reserve the remaining 30 for evaluation. Training samples undergo our proposed augmentation pipeline, hereafter denoted as GPDS-Aug, which is then used for training the generative model. All experiments are implemented in PyTorch and executed on an NVIDIA RTX A6000 GPU. Prior to training and testing, all images are uniformly resized to 224 × 224 × 3 pixels. We train with a batch size of 32 using the Adam optimizer. The learning rate is initialized at 5e-5 and annealed via a cosine schedule.

For quantitative benchmarking, we establish a multi-facet comparison: besides the conventional LGDR watermarking method \cite{ma2021local}, we evaluate three deep-learning approaches (HiDDeN \cite{zhu2018hidden}, CIN \cite{ma2022towards}, and WAM \cite{sander2024watermark}). The baseline methods are implemented according to their original configurations as described in the respective papers. For our method, \methodabbr{} embeds 48 bits of information.
Robustness is measured by extraction accuracy (Acc.). Visual fidelity is assessed using HWD \cite{pippi2023hwd}, FID \cite{heusel2017gans}, and KID \cite{binkowski2018demystifying}; HWD-tailored for handwriting-computes the Euclidean distance between the mean feature embeddings of genuine and synthesized signatures, sensitively reflecting stylistic deviations.

\subsection{Visualization Results} 
\begin{figure}[htbp]
    \centering
    \includegraphics[width=\columnwidth]{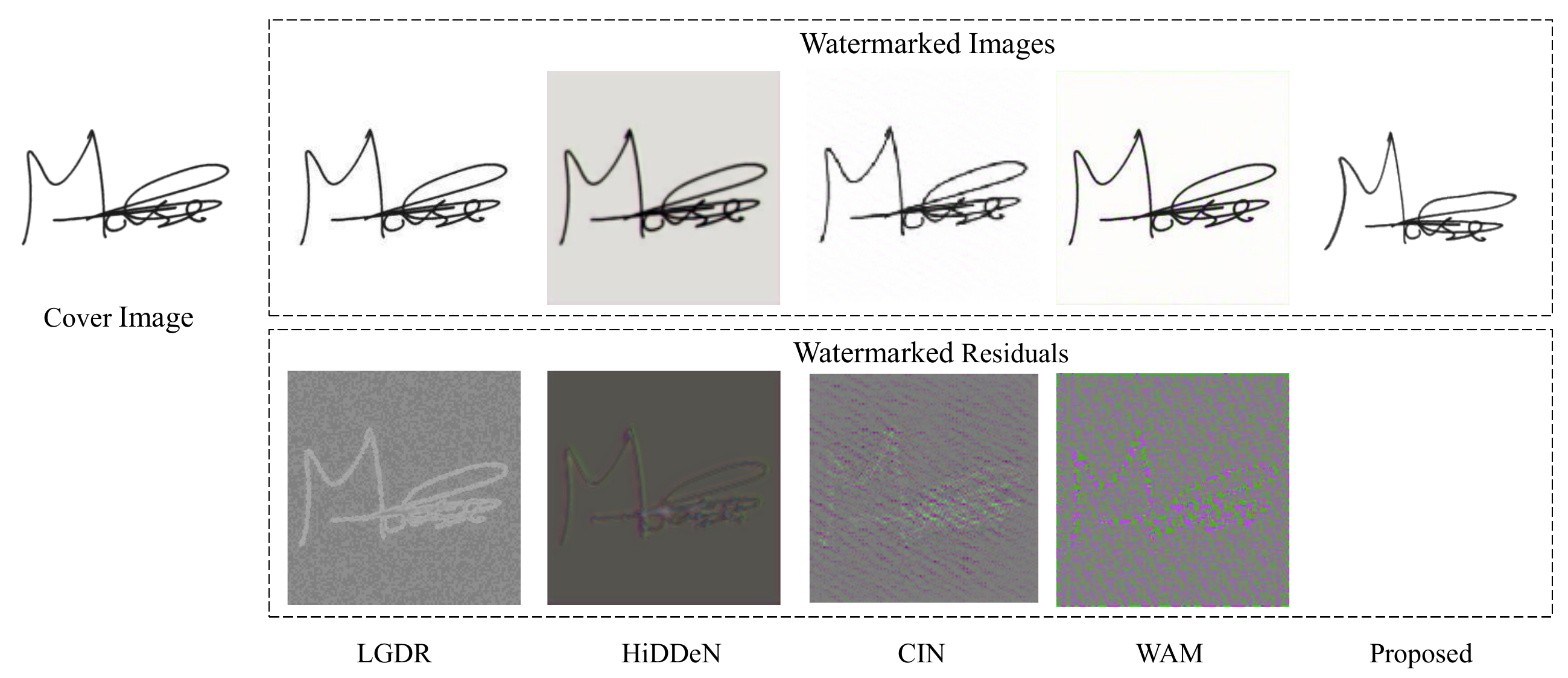}
    \caption{Visual examples of watermarked images \(I_{w}\) and residuals of the proposed scheme and the compared methods. left: cover image $I_0$. Residuals are calculated as $(I_{w}-I_{o}) \times 2+128$.}
    \label{fig:visualization_result}
\end{figure}

We qualitatively evaluate the visual quality of our scheme against four baseline methods: HiDDeN, WAM, LGDR, and CIN. As illustrated in Fig.\ref{fig:visualization_result}, HiDDeN introduces a pronounced background hue shift and blurs signature strokes due to pixel-level perturbations. WAM induces a global color drift and conspicuous green edge artifacts that are easily detected by the human eye. LGDR and CIN, while visually better, still exhibit compression artifacts that degrade fine-stroke details. Such distortions stem from their global embedding strategies and reliance on high-frequency image components. For instance, HiDDeN’s adversarial training enhances robustness on natural images, yet handwritten signatures constitute only roughly 14\% of pixels, compelling watermark embedding into extensive background regions and precipitating visual artifacts. Similarly, WAM's local embedding strategy, though designed to minimize background interference, is ill-suited for the sparse nature of signature data, leading to edge artifacts and color anomalies.

\begin{figure*}[t]
    \centering
    \includegraphics[width=\textwidth]{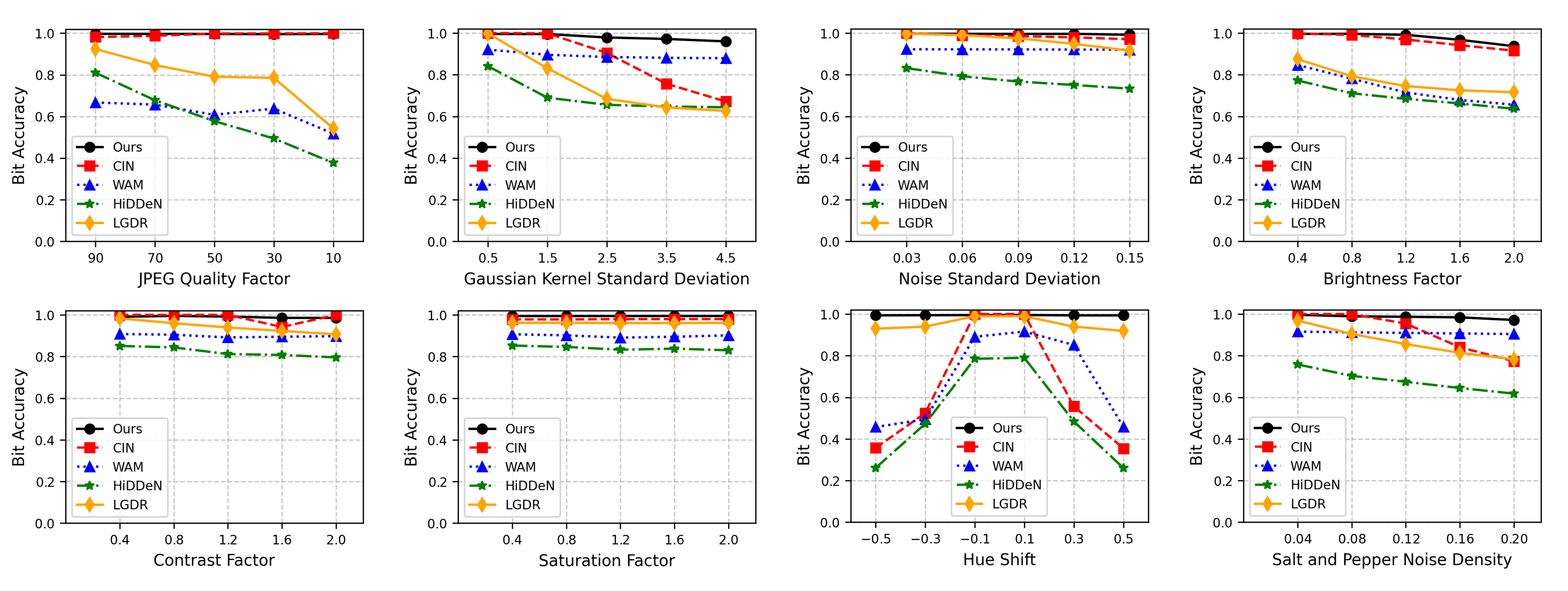}
    \caption{Robustness of our method versus competing watermarking schemes under common digital distortions.}
    \label{fig:ablation_studies}
\end{figure*}
In contrast, \methodabbr{} embeds watermark information by modulating subtle stylistic features of the signature(e.g., local strokes weight, turn curvature). This approach avoids perceptible distortions in the pixel domain, such as blurring, noise, or color shifts. The imperceptibility of our method is rooted in the natural tolerance of the Human Visual System (HVS) to minor variations in signatures. The stylistic appearance of any individual's signature—including its stroke morphology, ligature flow, and slant—inherently varies due to factors like the writing environment and the signer's emotional state. This intrinsic variability effectively masks the minute alterations introduced by the watermarking process, rendering them unnoticeable in practical applications. Since our scheme is based on a generative model rather than post-processing techniques used by the baselines, there is no conventional reference image for the watermarked signature, rendering residual computation infeasible.

\subsection{Robustness to Common Image Distortions}
We evaluate the robustness of \methodabbr{} against eight representative image distortions—JPEG compression, Gaussian blur, Gaussian noise, brightness variation, contrast variation, saturation change, hue shift, and Salt-and-Pepper noise. As illustrated in Fig. \ref{fig:ablation_studies}, our approach consistently outperforms baselines across all distortion types and severity levels, maintaining over 98\% extraction accuracy in most cases. This robustness arises from embedding watermark within the dynamic topological features of signatures rather than pixel intensities. 
In contrast, directly applying existing watermarking algorithms to handwritten data presents serious challenges. LGDR and HiDDeN suffer marked Acc drops under moderate to heavy JPEG compression, due to inherent conflicts between their high-frequency embedding strategies and JPEG’s DCT-based quantization. Similarly, existing methods show reduced performance under Gaussian blur, Gaussian noise, and Salt-and-Pepper noise. For instance, HiDDeN and CIN lose over 20\% accuracy under strong Gaussian blur. This may be because existing watermark algorithms, which assume embedding in high-frequency details of natural images, fail in such sparse data. The sparse feature distribution characteristics of the handwritten signature dataset (effective pixel ratio about 14\%) make existing schemes unable to maintain their robustness.

It is important to note that the algorithm's performance exhibits a degree of degradation when subjected to high-intensity brightness adjustments. Specifically, when the Brightness Factor reaches 2.0 (indicating a twofold increase in brightness), the Acc. decreases to 93.7\%. The underlying cause for this decline is overexposure, which results in the loss of strokes in certain signature samples with lighter handwriting. In extreme cases, this can lead to an image that is almost entirely white. Such an extreme scenario represents a critical state of signature image failure and is highly unlikely to occur in practical applications. Concurrently, 
comparative results indicate that existing methods degrade even more severely under such conditions.

\subsection{Robustness to Office Operations}
In digital office workflows, signature images are often subject to ROI-preserving cropping and background transparency, see Appendix A for details. To assess robustness under these common manipulations, we benchmarked \methodabbr{} against state-of-the-art baselines. As show in Table \ref{tab:real_use_expr}, \methodabbr{} consistently exceeds 99\% extraction accuracy for both operations, substantially outperforming competing methods. These results confirm the practical applicability of our scheme in real-world electronic document scenarios.
\begin{table}[t]
\setlength{\tabcolsep}{2pt}
\small
\centering
    \begin{tabular}{cccccc}
    \toprule
    \textbf{Metric} & \textbf{CIN}  & \textbf{LGDR} & \textbf{HiDDeN} & \textbf{WAM} & \textbf{Ours} \\
    \midrule
    BG. Transparency & 0.7195 & 0.5632 & 0.5225 & 0.4612 &  \textbf{0.9921} \\
    ROI-Preserving Crop & 0.7448 & 0.5814 & 0.5516 & 0.5936 & \textbf{0.9943} \\
    \bottomrule
    \end{tabular}
\caption{Comparison of Different Methods on Background Transparency and Cropping.}
\label{tab:real_use_expr}
\end{table}

\subsection{Robustness to Cross-Media Distortions}
To assess our method’s viability for cross-media signature authentication in real-world office settings, we conducted a three-phase physical experiment. First, we produced 205 watermarked signature images from the GPDS test set. Second, we printed them using two consumer printers (HP M233dw and M180n). Third, we re-digitized the prints with four devices—a Canon LiDE 300 scanner, an iPhone XR, an iPhone 15,and a vivoX100. To reduce device-specific variations, each print-and-scan trial was replicated three times. We report watermark extraction accuracy as our main metric. During mobile captures, phones were tripod-mounted and sheets fixed to a stand; all images were manually cropped and deskewed before decoding.

As shown in Table \ref{tab:wild_expr}, \methodabbr{} consistently exceeds 97\% accuracy across most device pairs. Crucially, it remains robust under challenging conditions-laser-induced ink spread and uneven lighting during handheld capture—matching the 97\% accuracy achieved with the uniformly lit LiDE scanner. Standard deviation of ±0.007 indicates stability across lighting and handling conditions. These findings demonstrate that our electronic signature scheme preserves the signature's authentication integrity even when the document is transferred from digital to physical medium (i.e., printed on paper). This capability overcomes limitations of conventional PKI digital signatures, which are restricted to electronic media, addressing the need for robust cross-media document authentication.

\begin{table}[t]
\centering
\setlength{\tabcolsep}{2pt}
    \begin{tabular}{ccccc}
    \toprule
     \textbf{Devices} & \textbf{iPhoneXR} & \textbf{iPhone15} & \textbf{vivoX100} & \textbf{Canon LiDE} \\
    \midrule
    HP M233dw   & 0.9662 & 0.9841 & 0.9724 & 0.9793 \\
    HP M180n    & 0.9712 & 0.9822 & 0.9796 & 0.9879 \\
    \bottomrule
    \end{tabular}
\caption{Extraction Acc. Across Various Printing and Scanning Device Configurations}
\label{tab:wild_expr}
\end{table}

\subsection{Effectiveness of Feature Decoupling}
To evaluate the feature decoupling mechanism, we conducted an ablation study with three variants: Full-Model (content + style), Style-Only (style without content), and Content-Only (content without style). Detailed visual results are provided in Appendix B, and quantitative evaluations are in Table \ref{tab:feature ablation}. Results show that high-fidelity reconstruction is achievable only when both features are jointly used. In contrast, single-feature models fail to reconstruct meaningful content, exhibiting severe semantic distortion. These findings highlight the necessity of combining content and style features for accurate signature image reconstruction.

\begin{table}[t]
\centering
\begin{tabular}{cccc}
\toprule
\textbf{Model} & \textbf{HWD $\downarrow$} & \textbf{FID $\downarrow$} & \textbf{KID $\downarrow$} \\
\midrule
Style-Only&1.7905 & 370.5272 & 0.4785 \\
Content-Only&1.7865 & 223.3530 & 0.2356 \\
Full-Model&\textbf{0.2047} & \textbf{40.8138} & \textbf{0.0418} \\
\bottomrule
\end{tabular}
\caption{Evaluation of the feature decoupling mechanism.}
\label{tab:feature ablation}
\end{table}

\begin{figure}[t]
    \centering
    \begin{minipage}[t]{\columnwidth}
        \centering
        \includegraphics[width=0.9\linewidth]{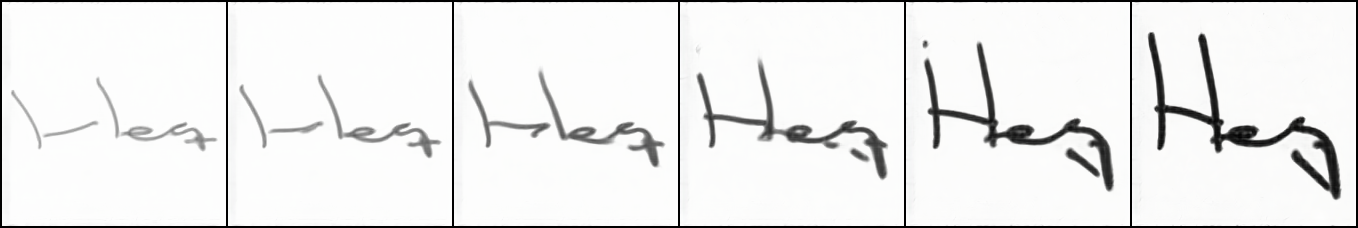}
    \end{minipage}
    
    \begin{minipage}[t]{\columnwidth}
        \centering
        \includegraphics[width=0.9\linewidth]{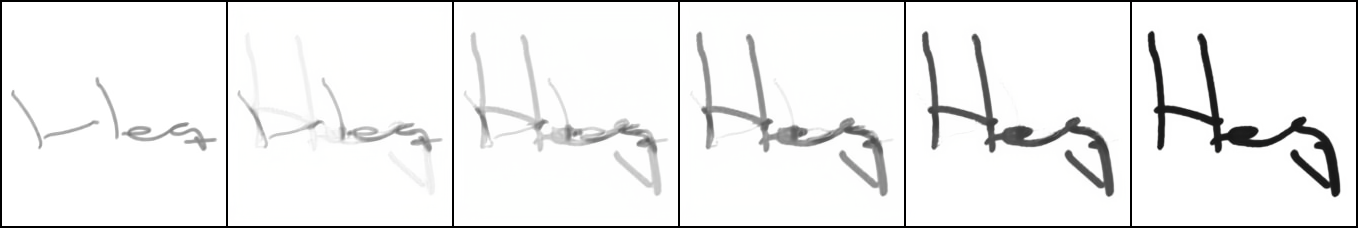}
    \end{minipage}
    \caption{Our method generates smooth transitions (top), while direct interpolation in the VAE latent space leads to superimposed artifacts from the two input images (bottom).}
    \label{fig:interpolate}
\end{figure}

Meanwhile, compared to directly using the latent variables of a Variational Autoencoder (VAE) to represent watermark messages, the proposed decoupling strategy models changes in the signature's style features at a finer granularity, achieving robust watermark embedding while preserving the natural characteristics of handwriting. As shown in Fig.~\ref{fig:interpolate}, entangled features in the latent spaces of a Variational Autoencoder (VAE) cause ghosting artifacts during feature interpolation. The proposed method effectively mitigates semantic conflicts and produces smooth, intermediate transition effects. Notably, the watermarked signatures are style-guided generations constrained within the distribution of genuine handwriting, ensuring realistic and authentic outputs. This capability further demonstrates that our generative model can freely sample within the feature space, enabling high-quality reconstruction of signature images and thereby providing ample capacity for watermark embedding; see Section D in Appendix.


\subsection{Effectiveness of VAE Finetuning}
We investigate the effect of fine-tuning a variational autoencoder (VAE) for handwritten signature reconstruction. Pretrained on generic datasets like LAION, existing VAEs poorly capture the fine-grained stroke patterns and high-frequency details of handwriting. To address these limitations, we fine-tune the VAE backbone from runwayml/stable-diffusion-v1-5 using domain-specific signature data. This adaptation improves latent space expressiveness, enabling more faithful detail preservation. We denote the pretrained model as PT-VAE and the fine-tuned variant as FT-VAE. 
As shown in Table \ref{tab:vae-performance}, the Reconstruction Handwritten Wasserstein Distance (rHWD) reduces from 0.1446 to 0.0344 and the Reconstruction Fréchet Inception Distance (rFID) on GPDS 1-150 drops from 5.76 to 1.63. Similar improvements are observed on the CEDAR, GPDS-Aug, and IAM datasets. Qualitative assessments further confirm enhanced continuity and fluency of strokes, surpassing the baseline in stylistic fidelity. These findings underscore the importance of VAE fine-tuning for high-fidelity generation tasks. Therefore, we recommend fine-tuning the VAE in advance to optimize the latent space for high-precision generation tasks, which is crucial for improving visual quality.


\begin{table}[t]
\centering
\setlength{\tabcolsep}{3pt}
\begin{tabular}{cccccc}
\toprule
\multirow{2}{*}{\textbf{Dataset}} & \multirow{2}{*}{\textbf{Model}} & \multicolumn{4}{c}{\textbf{Evaluation Metrics}} \\
\cmidrule(lr){3-6}
 &  & \textbf{PSNR ↑} & \textbf{SSIM ↑} & \textbf{rHWD ↓} & \textbf{rFID ↓} \\
\midrule
\multirow{2}{*}{GPDS 1-150} 
& PT & 32.54 & 0.9878 & 0.1446 & 5.76 \\
& FT & \textbf{44.37} & \textbf{0.9970} & \textbf{0.0344} & \textbf{1.63} \\
\midrule
\multirow{2}{*}{CEDAR} 
& PT & 33.24 & 0.9814 & 0.1471 & 5.44 \\
& FT & \textbf{43.12} & \textbf{0.9942} & \textbf{0.0513} & \textbf{1.97} \\
\midrule
\multirow{2}{*}{GPDS-Aug}
& PT & 33.35 & 0.9860 & 0.1564 & 6.18 \\
& FT & \textbf{41.01} & \textbf{0.9951} & \textbf{0.0414} & \textbf{1.72} \\
\midrule
\multirow{2}{*}{IAM} 
& PT & 34.42 & 0.9906 & 0.1611 & 5.02 \\
& FT & \textbf{40.68} & \textbf{0.9968} & \textbf{0.0713} & \textbf{0.80} \\
\bottomrule
\end{tabular}
\caption{Performance Comparison of Pre-trained VAE (PT) and Fine-tuned VAE (FT)}
\label{tab:vae-performance}
\end{table}


\section{Conclusion}
We propose a novel electronic signature framework that integrates watermark technology with generative models, imbuing static signatures with verifiability and document-binding capabilities. Comprehensive evaluations demonstrate its high resilience to common digital distortions and signature-specific degradations, even including print-scan process. Future work will target greater watermark capacity and improved visuals.

\section{Acknowledgments} 
This work was supported in part by the Natural Science Foundation of China under Grant 62402469, 62121002, 62472398, U2336206, by Fundamental Research Funds for the Central Universities under Grant WK2100000041.

\bibliography{aaai2026}

\end{document}